# Strategic Positioning in Tactical Scenario Planning

James M. Whitacre, Hussein A. Abbass, Ruhul Sarker
School of Information Technology, University of New South Wales at Australian Defence Force Academy, Northcott Drive, Canberra, Australia
61 2 626 88054

{j.whitacre,h.abbass,r.sarker}@adfa.edu.au

Axel Bender and Stephen Baker
Defence, Science and Technology Organisation, Edinburgh, Adelaide, Australia
61 8 825 96341

{axel.bender,steve.baker}@dsto.defence.gov.au

## ABSTRACT
Capability planning problems are pervasive throughout many areas of human interest with prominent examples found in defense and security. Planning provides a unique context for optimization that has not been explored in great detail and involves a number of interesting challenges which are distinct from traditional optimization research.

Planning problems demand solutions that can satisfy a number of competing objectives on multiple scales related to robustness, adaptiveness, risk, etc. The scenario method is a key approach for planning. Scenarios can be defined for long-term as well as short-term plans. This paper introduces computational scenario-based planning problems and proposes ways to accommodate strategic positioning within the tactical planning domain.

We demonstrate the methodology in a resource planning problem that is solved with a multi-objective evolutionary algorithm. Our discussion and results highlight the fact that scenario-based planning is naturally framed within a multi-objective setting. However, the conflicting objectives occur on different system levels rather than within a single system alone. This paper also contends that planning problems are of vital interest in many human endeavors and that Evolutionary Computation may be well positioned for this problem domain.

## Categories and Subject Descriptors
I.2.8 Problem Solving, Control Methods, and Search Heuristic methods

## 1. INTRODUCTION
In real-world planning problems, one must pose a question defining the purpose of the planning process, establish an appreciation of future environments, define the context(s) of these futures (scenarios), and then develop strategies/plans that are consistent with the organization's strategic goals. Planning is a continuous process and the dominant factor in planning problems is uncertainty.

Uncertainty is normally categorized in two types: branching points and deep uncertainty. The former represents the type of uncertainty we can expect; thus we can describe it with probabilistic techniques. The latter, however, is unanticipated uncertainty which causes discontinuity in the strategic space/environment. Deep uncertainty is normally a major factor in long-term strategic planning, where traditional trend analysis is more likely to fail in predicting the future. In tactical planning, the reliance on branching points is more sensible because meaningful probability distributions can be formed and get attached to the tactical scenarios in the scenario space.

Long-term planning can not work in isolation of tactical planning. The former defines the strategic position the organization needs to be taking to meet the challenges described by the strategic scenarios, while the latter needs to provide solutions that the organization can implement in the short run to meet the challenges imposed by the tactical scenarios. In reality, both types of planning need to be done hand-in-hand. Tactical decisions need to take into account strategic positioning so that short term decisions meet immediate threats while being in themselves steps towards meeting long-term threats.

This trade-off between short-term and long-term goals is a challenging task and should not be seen as a traditional multi-objective problem. One of the main challenges here is that the strategic position is normally described as a vision rather than a concrete set of quantifiable targets with unambiguous set of performance metrics. The second challenge is that the tactical planning problem is more about making decisions (because of its short term nature), while the strategic planning problem is more about understanding the future strategic environment to find ways to shape it. Therefore, using optimization techniques to model and find solutions for the tactical planning problem is acceptable. However, this raises another related challenge, namely that the potential conflict in the objective functions are on different levels of the system; that is, one objective is defined – normally qualitatively – on the strategic level while the other is defined more subtly on the tactical level. Interactive decision aiding tools are one way to separate the two problems, but not necessarily the preferred way because (1) only a few planners would understand the theoretical assumptions needed to drive such a system meaningfully; (2) the output could be insufficient due to the limited experience and the large bias the analyst may have; and (3) the tactical problem typically has some qualitative objectives which, in an interactive session, may be hard to distinguish from tactical objectives at the strategic level.

In general, planning problems are human-centric and planners are more comfortable with a methodology than an algorithm. As such, formulating the planning problem in an optimization model, and thinking that by solving this model the planning problem is solved, is misguided and may not be accepted by a planning analyst in the real world.

## 1.1 Understanding the Future

Making informed decisions when planning about the future generally involves either the application of prediction techniques or scenarios.

**Prediction Techniques**: Prediction techniques rely on historical data to generate a single expectation about the future. An often stated drawback is that the results from these techniques can be sensitive to the historical data sets available. When stated on its own, such an argument seems to imply that, given a large enough data set, a prediction technique could be a highly effective choice. More importantly however, prediction (and statistical learning in general) is fundamentally ill-equipped to express new emergent phenomena which have not been witnessed in previous events. Such phenomena are commonplace for instance in combat, biological coevolution, the global environment, technology markets and social systems. As a consequence, the applicability of a technique that relies on a generic model is limited to a constrained and relatively simple problem domain. Furthermore, if we are presented with an expansive range of possible futures, even the most probable future is still not that likely meaning that prediction is fundamentally ill-suited for such difficult planning problems. For more on the limitations of statistical predictive methods in planning problems, we refer the reader to [3].

**Scenarios**: To develop a robust plan for the future, it is necessary to understand the multitude of ways in which future events might unfold. To do this, most planning is traditionally carried out by groups of experts using human-derived scenarios. A scenario can be defined as "a postulated sequence of plausible events with some degree of internal coherence" [4]. The most important distinction between scenario-based methods and prediction is that scenarios allow us to capture a broad range of possibilities to what the future might hold. Unlike traditional optimization, we cannot optimize based on any single expected future making the problem domain interesting and unique.

Although most scenarios are presently derived by human experts, there are some well-recognized limitations to this approach including human fatigue, risks from normative thinking [4], and limits to the complexity of mental models. In recent years, there has been some notable progress in employing computational models (mainly agent-based systems) for modeling complex environments and it is becoming increasingly popular to develop hybrid scenario approaches that can exploit the unique capabilities of human-based and computational-based techniques [4][5][6][7][8]. As a consequence of these developments, we are for the first time in a favorable position to develop search heuristics for exploring computer-based simulation environments which in turn can be used as a support tool in planning for the future [3].

## 1.2 Why is EA a Good Choice?

From a practical standpoint, meta-heuristics are well suited for planning problems due to the design flexibility of these algorithms and the ability to quickly design solver tools that can mold to the needs of a client. From our experience, we have found that client needs for a planning problem's simulation environment can change quickly and repeatedly in this domain and we have found agent-based meta-heuristic search tools, if skillfully implemented, can keep up with these evolving needs. Despite the nice convergence properties of mathematical programming techniques, most commercial simulation packages that incorporate solver tools use meta-heuristics such as Evolutionary Algorithms (EA) [9] which we speculate is at least partly due to their flexibility and ease of implementation.

However, there may also be theoretical reasons why simulating evolution is a valuable way to think about addressing planning problems. As mentioned in the introduction, planning problems involve deep uncertainties about the future and there is no better exemplar of this than earth's environment. Despite its unpredictable dynamics and the dramatic changes that take place on local, regional, and global scales, there is one autonomous system which has proven capable of persisting within these difficult conditions, namely life. Instead of predicting the future, living systems appear to be designed with a natural robustness and adaptiveness which allow them to survive and indeed thrive in the face of great uncertainty.

Although robustness has been studied extensively in different optimization contexts, the role of adaptiveness in achieving robust behavior has received far less attention. Furthermore, the requirement of high robustness and adaptability should be of particular theoretical interest for evolutionary computation research since these very features are commonly used to distinguish natural evolution from optimization.

However, whether dealing with the planning problems considered here or dealing with natural environment, it is essential for systems/solutions to be robust to a broad range of different conditions and to be able to adapt to the unanticipated. These parallels are not present in other optimization contexts and we believe that this could provide a unique opportunity for the exploration of important theoretical topics within a domain of real practical interest.

## 1.3 Paper Outline

This paper is part of a broader set of investigations by this group into understanding planning problems with deep uncertainty and how heuristic search algorithms can be employed in this domain. Previous work has focused primarily on the development of a high fidelity simulation environment and agent-based evolutionary search methods for use in a military planning problem [10] [11].

In this paper, we focus squarely on the unique challenges related to tactical planning problems which also need to consider strategic positioning. Based on the proposed evaluation tools and other recommendations made in this paper, future work will develop a generic solver framework that is tailored to address the unique challenges in searching through the decision space (i.e. solution space) of scenario-based planning problems. Some preliminary sketches of this framework are outlined in the discussion.

To demonstrate the evaluation tools proposed in this paper, a tactical planning problem is presented where assets must be purchased to satisfy a range of short-term future capability needs. The problem definition and solver have been simplified so that this paper can focus on solver evaluation and performance analysis tools. For information on an agent-based multi-objective EA solver and military logistics simulation environment which these tools have been designed to support, we refer the reader to [10].

The rest of the paper is organized as follows. The next section describes scenario-based planning environments. Section 3 then outlines the metrics proposed in this paper for evaluating the strategic positioning of solutions. Section 4 presents a resource planning problem and a heuristic solver for generating solutions which are used to demonstrate the solution evaluation metrics. Results are presented in Section 5 with a discussion and conclusions completing the paper.

## 2. Scenario-Based Problem Solving

**Scenario Description:** There are many ways in which the future can unfold. The most popular way of representing uncertainty in the future is through scenarios. Scenarios provide a way of simplifying our view of the future by clustering possible future trajectories into distinguishable and meaningful groups. In particular, large-scale attributes about an environment which are meaningful to human decision makers are used to group futures into different scenarios. In a military context, a (simplistic but illustrative) set of scenarios might include peace-keeping, disaster recovery, and combat operations.

There are many details which are not outlined at the level of a scenario which are needed to fully specify a particular path to the future. Filling in these details involves specifying the structure of a model (i.e. model of the real environment) and specifying the initial conditions of that model. Defining these details can be thought of as instantiating the problem (e.g. see Figure 1), while running the model (i.e. simulating the dynamics of the real environment) allows us to generate an actual path to the future. Due to factors such as random external events, each simulation can take a unique path and result in different future conditions. The existence of multiple possible futures within a problem instance and multiple problem instances within a scenario is illustrated in Figure 1.

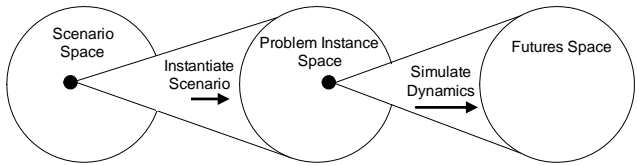

**Figure 1: Schematic overview of strategic scenarios with explicit distinctions made between scenarios, problem instances, and the simulation dynamics which generate possible futures.**

The topics that are relevant to a planning problem will depend on the timescale over which planning is taking place. Over short to moderate timescales (known as tactical and operational planning), the problem is viewed as one of strategic positioning based on relatively small time horizons such that emphasis is placed only on plan agility/flexibility and robustness. As the timescale is extended out to longer periods, a planning process also needs to account for its implementation strategy and practical challenges related to continuous learning.

Plan implementation, as the name implies, details out a path for how capabilities are built up and is typically constrained by budgetary limits and practical feasibility. Although these considerations are important, plan implementation becomes substantially more challenging when it attempts to address planning goals with unique time scales. For instance, the way in which a plan is implemented should not only address near term goals, but it should also be preparing for capabilities that will be important in the more distant future. Significantly adding to this challenge is the presence of *continuous learning*. With continuous learning, the plausibility of scenarios will evolve in real time. Thus the plan and its implementation must adapt with time to remain well positioned for success in both the short and the long term.

Tactical scenarios are different from the strategic ones in a few points: (1) deep uncertainty is less of an issue in a tactical scenario; (2) tactical scenarios are better defined and structured; it is easier to computer-simulate them; (3) short-term planning using tactical scenarios is more likely to generate suggestions for decisions to be taken by the decision maker, while long-term planning using strategic scenarios is more likely to define short-term desired objectives.

This paper addresses goals which are defined at a single time scale and as such, is more applicable to tactical and operational planning problems. The issues of plan implementation and continuous learning plans are not specifically dealt with in this paper, however these issues are addressed in our current research program.

For more information on scenarios and scenario-based planning problems, we refer the reader to the following reports [4] [12]. Although both of these reports discuss scenarios in the context of military planning, the applicability of the material is more general and may be of interest to the reader.

## 3. Options Analysis for Decision Support

### 3.1 Goals for Strategically Positioned Solutions

Here we describe the unique demands that are placed on a search tool when it is implemented in a planning context. The requirements listed here stem primarily from negotiations with our client however a number of reports from the RAND corporation are in general support of this view (e.g. [3][4][12][13]).

Planning problems require solutions that: i) have the ability to address a broad range of plausible futures, ii) can be quickly adapted to satisfy conditions falling outside the current set of capabilities, and iii) are robust in the sense that a solution's viability is not fragile to plausible damaging (or unanticipated) events.

This implies that a solution should NOT be optimized to solve a single problem instance of a single scenario. It should instead be well-positioned to address a number of possible futures (expected performance), have feasible and low cost changes available that can result in relevant capability changes (adaptiveness), and be able to remain a viable option in the face of significant unanticipated changes to problem conditions (high robustness).

While this resembles robust optimization, this form of optimization is normally done using a single criterion. The problem we are addressing here is robust optimization under multiple conflicting criteria. This raises an interesting challenge that there can be multiple robust solutions, each of which has a different degree of robustness against the different criteria. Furthermore, robust optimization does not directly address the issue of adaptation. Adaptation becomes particularly relevant in planning problems where solutions are valued based on their capacity to favorably respond to significant changes in environmental conditions.

Our approach to developing computational tools for tactical planning problems involves the steps outlined below.

1. Identify the strategic position based on a strategic planning exercise.
2. Develop and model tactical scenarios.
3. Identify an acceptable set of tactical goals.
4. Develop a solver using search tools with significant algorithm design flexibility such as multi-objective Evolutionary Algorithms and hybrid search techniques
5. Generate a solution space and assess each solution relative to the developed scenarios.
6. Choose a solution based on trade-offs of performance in the tactical scenarios and the strategic position.

We will assume in this paper that the strategic position is known and is described as a vision in a qualitative manner. We start by talking briefly about performance evaluation within an individual scenario.

## 3.2 Evaluating the Strategic Position of Solutions

### 3.2.1 Performance within a Scenario

Evaluating a solution's performance in a tactical scenario is a non-trivial task because a scenario can be instantiated (i.e. a problem instance created) in a number of different ways, and the simulation of each problem instance is free to evolve to a number of different possible futures (see Figure 1). Every one of these possible futures may evaluate the effectiveness of a solution differently. To understand how a solution performs within a scenario, it is therefore necessary to evaluate the solution within a number of possible futures. Thus for each scenario in the scenario space, a solution will be given a set of objective function evaluations.

Comparing the performance of different solutions therefore requires to compare sets of evaluation data (e.g. by using statistical tests) or to distil a single metric from a solution's evaluation data and then compare the solution with others. This task can be further complicated by the possibility of multiple objectives (as seen in the case study in Section 4) or the need for aspirational targets for objectives (e.g. as is used in goal programming). Regardless of the procedure used, this is a point where a decision maker would be injecting significant bias into the performance calculations. Since the best approach to measuring solution performance within a scenario generally depends on the planning problem being investigated, we intentionally do not specify the details of calculation steps in this section but assume that the evaluation data sets associated with each solution in each scenario have been aggregated in some meaningful way. In Section 4 we specify this aggregation step in the context of a case study.

Throughout the remainder of the paper, we refer to the performance of the $i^{th}$ solution $s_i$ within scenario $j$ as $F_j(s_i)$. As shown in (1), we simply assume that the performance measure involves some function which compares $s_i$ to some set $S^*$ of other solutions in a scenario $j$. A particular example of (1) is given later in the case study in Section 4.

$$F_j(s_i) = fn(s_i, S^*|j) \qquad (1)$$

### 3.2.2 Solution Quality and Robustness

The first step in understanding the quality of a solution is to measure its robustness in the face of future uncertainty. A calculation of a solution's robustness should indicate the solution's ability to succeed in a range of plausible scenarios. Assuming that we have a single performance metric like (1) and given an aspirational target $F_j^{aspire}$ which represents a lower threshold for defining success within a scenario $j$, the overall robustness for a solution can then be estimated in (2) as the percentage of scenarios that a solution can perform successfully. Here we care little about overall expected performance (such as the sum of $F_j(s_i)$ over all scenarios) and instead focus on attaining a certain performance level across as many plausible futures as possible. In tactical planning, it is acceptable to use probabilities to describe how frequently a scenario may occur relative to other scenarios that are within the set of $Q$ scenarios under consideration. This is accommodated in (2) through the use of the probability density function $P(j)$.

$$F(s_i) = \sum_{j=1}^{Q} \left( F_j(s_i) \geq F_j^{aspire} \right) * P(j) \qquad (2)$$

### 3.2.3 Risk

Risk is defined here as vulnerabilities in the system based on a biased decision (i.e. the choice of one solution that satisfies certain scenarios and not others). This risk represents the possibility of solution failure in a scenario. This form of risk is definable by characterizing the distribution of evaluation data for a solution within a scenario. As an example, one might define a minimum threshold for acceptable objective function values and the probability of crossing this threshold would indicate a key risk of unacceptable performance (i.e. failure).

An important goal in planning problems is to avert disastrous situations which we account for by minimizing the risk of failure within a scenario as defined in (3). Assuming some definition of failure within a future environment, we can calculate the expectation of failure $Failed_j(s_i)$ by testing the solution $s_i$ on many simulations within a scenario. The total expected risk of failure can then be calculated by summing this over all scenarios and again adjusting by the probability of each scenario.

$$Risk(s_i) = \sum_{j}^{Q} Failed_j(s_i) * P(j) \qquad (3)$$

### 3.2.4 Adaptiveness

As implied in Figure 2, a solution that is suited for one scenario may not be suitable for another. Since we do not know in advance which of the stated futures will actually eventuate (if any), we must position ourselves in a manner that best prepares us for a range of possibilities.

Adaptiveness in planning problems means that a solution is well positioned to make feasible changes that improve a solution's performance/capabilities within a particular future environment. The feasibility of change is largely constrained by the timescale needed to implement such changes. In the context of tactical planning, this timescale is too short to allow for major acquisitions. Instead, feasible adaptations consist of organizational restructuring, reallocation of assets, and retooling operations where existing capabilities are slightly modified to take on a task they may not have been specifically designed for.

If we assume that the decision space (i.e. solution space) can represent such morphological changes then we can define in (4) a solution $s_i$'s adaptiveness (or, more precisely, the cost of adaptiveness) as the aggregated costs incurred during (localized) movements from one position in decision space to another. This movement corresponds to modifying a solution so that it has the same capabilities as the most desirable solution $s_j^{Best}$ that can be reached through feasible modifications to $s_i$ in each scenario $j$.

$$Adapt(s_i) = \sum_{j}^{Q} Cost(s_i \rightarrow s_j^{Best}) * P(j) \quad (4)$$

It is worth noting that there is an implicit tradeoff that must be accounted for when defining $s_j^{Best}$. In particular, there is a tradeoff between the amount of adaptive improvements made and the cost of such adaptations. How this balance is chosen is actually an important feature of a solution's strategic positioning which suggests that part of a solution's definition should include how this balance is selected. For example, this may require the solution to include parameters which specify bounds on maximum adaptive costs, bounds on minimum performance, or ratios of minimally acceptable levels of performance gain per unit cost.

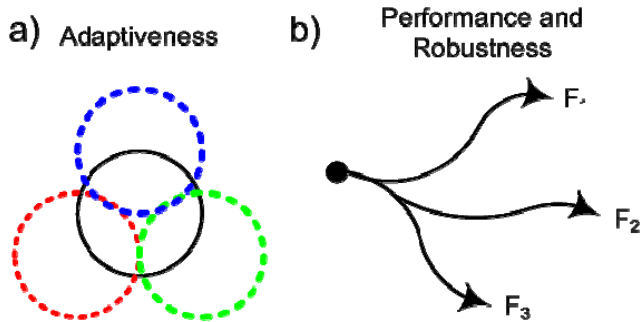

**Figure 2:** Evaluating solutions in a scenario-based planning environment. a) The black circle represents a strategically positioned solution while feasible adaptive options for different scenarios are shown as colored, dashed circles. Portions of a colored circle not included in the black circle represent the cost of adaptation to a particular environment. b) A solution (black dot) is evaluated within a number of possible future environments.

In this section we have proposed three generic metrics (2), (3) and (4) for tactical scenario planning. These metrics could in turn be used explicitly as new high-level objectives to a planning problem. To do this however requires solutions to represent more than just points in decision space but to also include feasible and meaningful adaptive options.

Alternatively, these metrics could simply be used to better understand the properties of a set of final solutions that have been evolved based on some other objectives such as the performance metric given in (1). This is the approach we take in this work.

In either case, a decision maker is likely to be better informed about their available options if these options are evaluated and presented based on the expected robustness (2), risk (3) and adaptiveness (4). These goals are conflicting which would imply that strategic planning is by its very nature a multi-objective problem.

The next section describes a set of experiments which have been devised to demonstrate the evaluation metrics proposed in this section.

## 4. Case Study
### 4.1 Resource Planning Problem

In the resource planning problem used in our experiments, there are $n$ asset types with the number of assets of type $i$ given by $x_i$, $i \in n$, and the unit cost of asset type $i$ given by $c_i$. One unit of asset type $i$ has the capacity to satisfy a given capability demand of type $k$ by an amount $w_{ik}$. The parameter $w_{ik}$ takes a value of zero when an asset is not suitable for a particular capability demand.

Within a given problem's time horizon $T_h$, we have a set of $I_t$ equally spaced discrete time points during which a set of capability demands occur. At each of these points in time, demands can occur for each of the $m$ types of capability demands with the size of the demands denoted by $d_k$, $k \in m$.

For each simulation (i.e. path to the future), the capability demands $d_k$ for every decision point and each demand type $k$ is sampled from a normal distribution with mean $\mu_k$, and standard deviation $\sigma_k$. For all experiments we set $I_t = 10$, $n = 5$, $m = 4$. Additional asset-related parameters settings are given in Table 1.

**Table 1: Asset parameter settings**

| Asset | Cost ($c_i$) | $w_{i,1}$ | $w_{i,2}$ | $w_{i,3}$ | $w_{i,4}$ |
|---|---|---|---|---|---|
| 1 | 1 | 3 | 3 | 3 | 3 |
| 2 | 1 | 1 | 6 | 5 | 0 |
| 3 | 1 | 0 | 0 | 6 | 6 |
| 4 | 1 | 10 | 0 | 0 | 2 |
| 5 | 1 | 0 | 4 | 4 | 4 |

### 4.2 Scenario Generation

For the resource planning problem, different scenarios are characterized by different settings for $\mu_k$ and $\sigma_k$ which allows for a hypothetical investigation of different capability requirements in the future. The settings tested in these experiments are given in Table 2.

Uncertainty in the overall scale of future capability requirements is used as a distinguishing factor between different problem instances within a scenario. In particular, for each problem instance a scaling factor $\beta$ is uniformly sampled from the interval [1,10] which is then multiplied to all values $d_k$ that are generated for a problem instance. Simulation dynamics within a problem instance simply involve the stochastic sampling of $d_k$ values which is constrained by the distributions set forth in each scenario and given in Table 2. It is worth noting that in our case study the scenario definition and simulation environment has been simplified and does not capture all of the properties that one would expect to observe in a real-world planning problem.

**Table 2: Scenario settings for the resource planning problem**

| Scenario | $\mu_k/\sigma_k$ (k=1) | (k=2) | (k=3) | (k=4) | P(j) |
|---|---|---|---|---|---|
| 1 | 2/1 | 2/3 | 3/4 | 3/2 | 0.25 |
| 2 | 10/4 | 6/3 | 6/2 | 7/2 | 0.25 |
| 3 | 0/1 | 10/1 | 9/4 | 5/4 | 0.25 |
| 4 | 4/2 | 6/2 | 6/3 | 5/3 | 0.25 |

### 4.3 Solver Description

The solver we implement has two components: an asset assignment heuristic for combining assets with capability demands during a simulation and an Evolutionary Algorithm (EA) for determining the amounts of each asset to invest in.

**Asset Assignment Heuristic Pseudocode:**

1. gather all assets that are available in the given time window and which are able to handle the given capability demand
2. select assets that have the lowest cost per unit of capability demand satisfaction
3. if an excess of asset options exist, select at random from the set
4. If capability demand was larger than capacity of assets selected, go to step 1 with capability demand sizes reduced by the amount satisfied in the last loop iteration

**EA Description:** The EA has $n$ integer value genes where the $x_i$ gene represents the amount of the $i^{th}$ resource that is available for use in the resource assignment problem.

Each solution is evaluated by two objectives. The first objective is given in (5) and is simply the total cost of investing in the resources specified by a solution's genes. The second objective is a measure of the performance of a solution within a single scenario $j$. In particular, we say a solution $s_i$ has succeeded in any simulated future when the available resources are sufficient to satisfy all of the demands. For the $h^{th}$ simulated future $h_j$ in $j$, we define $Success(s_i|h_j)$ as a Boolean function indicating whether a solution succeeded. The second objective in (6) is then given by the rate of success within a scenario, i.e. the average success of a solution over the $r$ futures in scenario $j$. It is worth noting that (6) depends on both the solution and scenario being considered while (5) only depends on the solution being considered.

$$Cost = \sum_{i=1}^{n} c_i x_i \quad (5)$$

$$\% \, Success = \frac{1}{r} \sum_{h_j=1}^{r} Success(s_i|h_j) \quad (6)$$

The EA is a simple steady state design where the parent population is a non-dominated set and parents are selected at random to create new offspring. Offspring are generated by using uniform crossover followed by Gaussian mutation with probability $2/n$ per gene and standard deviation $\sigma = 0.1$. The initial population size $\mu$ is set to 20 and initially some parents will likely be dominated by others. The population updating is implemented such that non-dominated offspring replace any member of the parent population that they dominate or that are dominated by other members. If no such members exist then the offspring is added and the population grows by one. Pseudocode for the algorithm is given below.

**Pseudocode for EA**

Randomly initialize parent population $\mu$
Do
    Select Parents $P_1$, $P_2$ from $\mu$ at random
    Generate Child $C_1$ (recombine, mutate)
    Evaluate $C_1$ and calc non-dom rank (Rank)
    If ( $\forall$ Rank($P_3$) > Rank($C_1$) | $P_3 \in \mu$ )
        Replace $P_3$ with $C_1$ in $\mu$
    Else If (Rank($C_1$)=1)
        Add $C_1$ to $\mu$
    End If
Loop Until Stopping Criteria

The EA finishes after 2000 evaluations. To calculate (6), 10 problem instances are generated, each of which are simulated 10 times, meaning that 100 simulations ($r$=100) take place to evaluate a single solution within a scenario.

### 4.4 Instantiating Strategic Positioning Metrics

Once the non-dominated sets of solutions are generated for each scenario of the resource planning problem, these final solutions are then evaluated based on their strategic positioning in decision space using the metrics described in the last section. Specific details needed to evaluate the strategic positioning of solutions are given below.

**Robustness**: Calculating solution robustness as defined in (2) requires the definition of a single measure of solution performance within a scenario of the form given in (1). This is accomplished by weighting and combining the two objectives given for the resource planning problem which are defined in (5) and (6). In particular, we take the final set of non-dominated solutions for a scenario, transform the objective functions to each be bounded between [0,1] with optimal values (highest success and lowest cost) of 1, and finally aggregate the objectives where (5) is multiplied by a weight of 0.3 and (6) is multiplied by 0.7. Based on this calculation of (1), which is also bounded between [0,1], we finally specify $F_j^{aspire} \equiv F^{aspire} \equiv 0.8$ for all scenarios $j$.

**Risk**: To calculate risk, we define failure in a scenario when a solution has a success rate (6) of less than 60%.

**Cost of Adaptation**: We assume that $s_j^{Best}$ is the same for all solutions for a given scenario and is simply the solution with the best performance in the scenario (in other words, we neglect feasibility constraints to adaptation). The cost of these adaptations is calculated as the sum of all assets needed to reach $s_j^{Best}$ multiplied by the respective cost per unit of asset $c_i$.

## 5. Results

To illustrate the difference between scenario-based planning problems and traditional optimization domains, Figure 3 shows solutions that are evaluated on each of the four different scenarios. To be clear, these solutions are the non-dominated sets from each scenario which are then evaluated on all other scenarios.

These results highlight the fact that, unlike other multi-objective problem types, each solution has multiple evaluations (for each scenario) which can make it more difficult to understand the tradeoffs between different solutions. We note that the number of data points is directly proportional to the number of scenarios making this presentation of results also unreasonable if we want to explore a huge number of plausible scenarios as has been advocated in the literature [4].

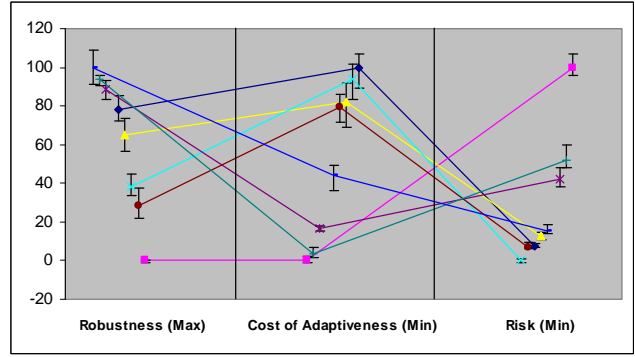

**Figure 4: Non-dominated solutions from maximizing (2) and minimizing (3) and (4). Metrics have been normalized over [0,100] for visualization purposes. Error bars provide a sensitivity analysis to the setting of scenario priorities $P(j)$.**

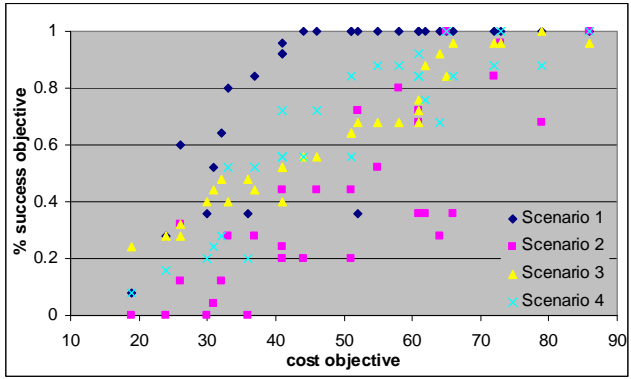

**Figure 3: Objective space for individual scenarios. Each solution has a single static cost (5) but different success rates (6) for each of the four scenarios. Objective function data is shown as a different data series for each scenario.**

What is also missing in Figure 3 and what only becomes relevant in planning problems is the issue of strategic positioning of solutions. Strategic positioning requires solutions that are robust within many different scenarios (which might be possible to ascertain from the results in Figure 3), however it also requires an understanding of what adaptations are possible and what risks are associated with a solution. The risk and adaptation measures we have defined in Section 3 do not necessarily capture all aspects of strategic positioning; however, they are reasonable surrogates and sufficient to illustrate the need for quantitative measures of strategic positioning.

To understand the robustness, risk and adaptiveness associated with found solutions, we present their evaluations against these attributes in Figure 4. Here we only present those solutions which are non-dominated with respect to the three metrics. As can be seen, there is clearly a tradeoff between these strategic positioning goals.

**Sensitivity to Bias**: It is also important to understand how strategic positioning is influenced by biases in the calculation steps which arise from the use of probabilities for different scenarios $P(j)$ and the use of weights for aggregating the objectives (5) and (6). To account for this, we show error bars in Figure 4 which indicate the sensitivity of each of the metrics to variations of probability values. Here the error bars are determined by the 1st and 3rd quartiles for metric values when $P(j)$ settings are jointly sampled over a Gaussian with standard deviation 0.1. A similar procedure is conducted to understand sensitivity to the objective weights (used to define (1)) and is presented in Figure 5.

It is worth noting that even in this simple resource planning problem, we find interesting sensitivities to the setting of objective function weights and to scenario probabilities. In Figure 4 we see that different solutions have different sensitivities to changes in scenario probabilities although on average this does not tend to impact preference rankings. However as we vary objective weights in Figure 5, we find the results are more sensitive to this bias and in particular, cause most solutions to have an added cost to adaptation.

**Selecting a plan to implement:** Although the high-level objectives of robustness, adaptiveness, and risk were found to reduce the number of non-dominated solutions compared to the original set presented in Figure 3, a decision maker is still left with some decisions for selecting a final solution to implement. Establishing minimally acceptable levels for robustness, the cost of adaptiveness, and risk could quickly reduce the non-dominated set to a much smaller number. In practice, qualitative objectives would also be applied at this point which may also eliminate many of the non-dominated solutions from consideration.

The final set of solutions should help inform the tactical decision making process, however it should also be informative at the strategic level. These solutions could provide strategic planners with useful information regarding the tradeoffs and bounds on robustness, risk, and adaptive capability. Planners are also likely to be interested in obtaining more detailed information about particular sources of risk, conditions for robustness, and modes of adaptiveness which would require a thorough analysis of performance across the scenario space as done for instance in [4].

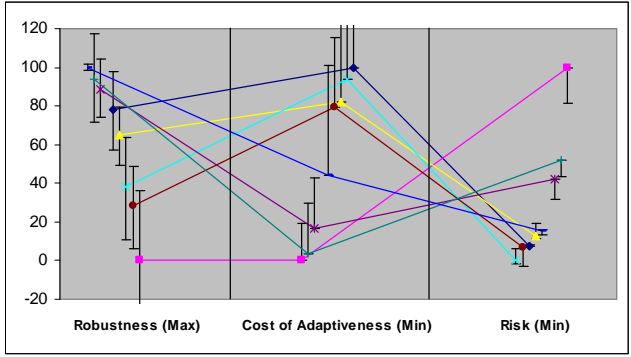

**Figure 5: Results repeated from Figure 4 except with error bars generated by varying the weights for objectives (5) and (6).**

## 6. Future work

Given a better understanding of what it means for a solution to be strategically positioned in the decision space, it is important to then think about how this changes our definition of a solution and how we think about searching through such a space.

The strategic position of solutions requires an understanding of how a solution performs in multiple plausible environments as a consequence of the uncertainty about the future. However, we also stress that planning is a fluid environment which sometimes requires adaptive capabilities. Once adaptation is explicitly accounted for, the standard notion of static solution points in decision space becomes noticeably inadequate for our needs. Instead it is more useful to think of solutions in terms of both their current state and a set of feasible future changes.

In future work, we intend to develop a "clustered coevolutionary" framework for solving planning problems where individual solutions co-evolve based on: i) complementary capability development across the scenario space and ii) feasible morphological relationships in the decision space. Solutions to the overall planning problem would then represent clusters in decision space where individuals are effective in their own right in certain conditions but are able to lower their risk and improve overall effectiveness through their interaction/cooperation with complementary morphological neighbors. By driving the evolution of such a system using the strategic positioning metrics defined in this paper, we hope to develop a robust solver tool for supporting decision makers in planning for the future.

## 7. Conclusions

In this paper we have described scenario-based tactical planning problems. We presented several unique requirements in evaluating solutions based on their strategic positioning in decision space which are distinct from optimization to some known objective function(s). We claim that evaluating the strategic positioning of a solution requires an understanding of a solution's robustness, risk, and adaptability and we proposed metrics for calculating these properties. The key differentiator between traditional multi-objective optimization and the multi-objective problem we addressed in this paper is that in the former, all objectives are defined on the system level, while in the latter, we address a problem on multiple scales.

We also note that this is a problem domain that requires tools which are flexible and can be quickly implemented making meta-heuristics such as EAs one of the few practical tools available for supporting computational studies of scenario-based planning problems.